\documentstyle[aclap]{article}

\title{Fast Statistical Parsing of Noun Phrases \\ for Document Indexing }

\author{Chengxiang Zhai\\
Laboratory for Computational Linguistics\\
Carnegie Mellon University\\
Pittsburgh, PA 15213\\
\{cz25@andrew.cmu.edu\} }

\begin{document}

\maketitle
\begin{abstract}

Information Retrieval (IR) is an important application area of Natural
Language Processing (NLP) where one encounters the genuine challenge
of processing large quantities of unrestricted natural language text.
While much effort has been made to apply NLP techniques to IR, very few
NLP techniques have  been evaluated on a document collection larger
than several megabytes. Many NLP techniques are simply not efficient 
enough, and not robust enough, to handle a large amount of text. 
This paper proposes a new probabilistic model for noun phrase parsing,
and reports on the application of such a parsing technique to 
enhance document indexing. The effectiveness of 
using syntactic phrases provided by the parser to supplement single
words for indexing is
evaluated with a 250 megabytes document collection. The experiment's
results show that supplementing single words with
 syntactic phrases for indexing consistently
and significantly improves retrieval performance.

\end{abstract}

\section{Introduction}

Information Retrieval (IR) is an increasingly important application area of
Natural Language Processing (NLP). An IR task can be described as to
find, from a given document collection, a subset of documents whose content is
 relevant to the information need of a user as expressed by a query. 
As the documents and query are often natural language texts, an IR task
can usually be regarded as a special NLP task, where the document text and the
query text need to be processed  in order to 
judge the relevancy. A general strategy followed by most IR systems is
to transform documents and the query into certain level of representation.
A query representation can then be compared with a document
representation  to decide if the document is relevant to the query. In
practice, the 
level of representation in an IR system is quite ``shallow'' --- often merely
a set of word-like strings, or indexing terms. The process to extract
indexing terms from each document in the collection is called indexing.
A query is often subject to similar processing, and the relevancy is
judged based on the matching of query terms and document terms. In most
systems, weights are assigned to terms to indicate how well they can be
used to discriminate relevant documents from irrelevant ones.

The challenge in applying NLP to IR is to deal with a large amount
of unrestricted natural language text. The NLP techniques used must be
very efficient and robust, since the amount of text in the databases accessed
is typically measured in gigabytes. In the past, 
NLP techniques of different levels, including morphological,
syntactic/semantic, and discourse processing, were exploited to
enhance retrieval \cite{Smeaton,Lewis-ACM}, but were rarely evaluated using 
 collections of documents larger than several megabytes.
Many NLP techniques are simply 
not efficient enough or are too labor-intensive to successfully handle a large 
size document set. However, there are some exceptions. Evans et al. 
used selective NLP techniques, that are especially robust and efficient,
for indexing \cite{CLARIT-RIAO-91}. Strzalkowski  reported a fast and
robust parser called TTP in \cite{NYU-nlp,NYU-ACL}. These NLP techniques
have been successfully 
used to process quite large collections, as shown in a series of  TREC
conference reports by  the
CLARIT$^{TM}$\footnote{CLARIT is a registered trademark of CLARITECH
Corporation.} system group and the New York University (later GE/NYU)
group (cf., for example, \cite{CLARIT-TREC2,CLARIT-TREC4}, and
\cite{NYU-TREC,NYU-TREC4})
These research efforts 
 demonstrated the feasibility of using selective
NLP to handle large
collections.  A special NLP track emphasizing the evaluation of NLP
techniques for IR is currently held in the context of TREC \cite{TREC5}. 
 
In this paper, a
fast probabilistic noun phrase parser is described. The parser can be
exploited  to 
automatically extract syntactic phrases from  a large amount of 
documents for indexing.
A 250-megabyte document set\footnote{the Wall Street Journal
database in
Tipster Disk2 \cite{TREC5}} is used to evaluate the effectiveness of
indexing using the phrases extracted by the parser.
The experiment's results show that using syntactic phrases to
supplement single words for indexing improves the retrieval
performance significantly. This is quite encouraging compared to earlier
experiments on phrase indexing.
The noun phrase parser provides the possibility of combining different kinds of
phrases with single words.   

The rest of the paper is organized as follows. Section~\ref{indexing} discusses
document indexing, and argues for the rationality of using syntactic
phrases for indexing; Section~\ref{em} describes the fast noun  phrase
parser that we use to extract candidate phrases; Section~\ref{exp} describes
how we use a commercial IR system to perform the desired  experiments;
Section~\ref{res} reports and discusses the experiment results; 
 Section~\ref{con} summarizes the conclusions.

\section{Phrases for Document Indexing}
\label{indexing}
In most current IR systems, documents are primarily indexed by  
single words, 
sometimes supplemented by phrases obtained with  statistical approaches, such
as frequency counting of adjacent word pairs. However, single words are
often ambiguous and not specific enough for accurate discrimination of 
documents. For example, only using the word ``{\em bank}'' and ``{\em
terminology}'' 
for indexing is not enough to distinguish ``{\em bank terminology}'' from
``{\em terminology  bank}''. More specific indexing units are needed.  
Syntactic phrases (i.e., phrases with certain syntactic relations) are
almost always more specific than single words and thus are intuitively
attractive for indexing. For example, if ``{\em bank terminology}''
occurs in the document,
then, we can use the phrase ``{\em bank terminology}'' as an additional
unit to supplement
the single words ``{\em bank}'' and ``{\em terminology}'' for indexing.
In this way,
a query with ``{\em terminology bank}'' will match better with the
document than
one with ``{\em bank terminology}'', since the indexing phrase ``{\em
bank terminology}''
provides extra discrimination.

Despite the intuitive rationality of using phrases for indexing, 
syntactic phrases have  been reported to show
no significant improvement of retrieval performance
\cite{Lewis,Belkin,Fagan}. Moreover Fagan \cite{Fagan}  found  
that syntactic phrases are not superior to simple statistical phrases. 
Lewis discussed why the syntactic phrase indexing has not worked
and concluded that the problems with syntactic phrases are for the most
part statistical \cite{Lewis}. Indeed, many (perhaps most) syntactic
phrases  have very 
low frequency and tend to be over-weighted by the normal weighting method.
However, the size of the  collection used in these early experiments is
relatively
small. We want to see if a much larger size of collection will make a
difference. It is possible that a larger document  collection might
increase the frequency of most 
phrases, and thus alleviate the problem of low frequency.

We only consider noun phrases and the sub-phrases derived from them.
Specifically, we
want to obtain the full modification structure of each noun phrase in
the documents and query. From the viewpoint of NLP, the task is  noun
phrase parsing (i.e., the analysis of noun phrase structure). When the
phrases are used only to supplement, not replace, the single words for
indexing,  some parsing errors may be tolerable. This
means that the penalty for a parsing error may not be significant. The
challenge, however, is to be able to parse gigabytes of text in
practically feasible time and as accurately as possible. The previous
work taking on this
challenge includes
\cite{CLARIT-RIAO-91,CLARIT-TREC4,ACL,NYU-TREC2,NYU-TREC3} among others.
Evans et al. 
exploited the ``attestedness'' of subphrases to partially reveal 
the structure of long noun phrases \cite{CLARIT-RIAO-91,CLARIT-TREC4}.
Strzalkowski et al. 
adopted a fast
Tagged Text Parser (TTP) to extract head modifier pairs including those
in a noun phrase \cite{NYU-nlp,NYU-ACL,NYU-TREC2,NYU-TREC3}. In
\cite{NYU-TREC3}, the structure of a noun phrase is disambiguated 
based on certain 
statistical heuristics, but there seems to be  no effort to assign a
full structure to every noun phrase. Furthermore, manual effort is
needed in constructing
grammar rules. Thus, the approach in \cite{NYU-TREC3} does not address
the special need of scalability and robustness along with speed. 
Evans and Zhai explored a hybrid noun phrase analysis method and used 
a quite rich set of phrases for document indexing \cite{ACL}. The
indexing method
was evaluated using the Associated Press 
newswire 89 (AP89) database in Tipster Disk1, and  a general improvement
of retrieval performance over the indexing with single words and full
noun phrases was reported. However, the phrase extraction system as
reported in \cite{ACL} is still not fast enough to deal with document
collections measured by
gigabytes.\footnote{It was reported to take about 3.5 hours to process
20 MB documents}

We propose here a probabilistic model of noun phrase parsing. A fast
statistical noun phrase parser has been developed based on the 
probabilistic model. 
The parser works fast and can be scaled up to
parse gigabytes text within acceptable time.\footnote{With a 133MH DEC
alpha workstation, it is estimated to parse at a speed of 4
hours/gigabyte-text or 8 hours/gigabyte-nps, after 20 hours of training
with 1 gigabyte text} Our
goal is to generate different kinds of candidate syntactic phrases
from the structure of a noun phrase so that  the effectiveness of
different combinations of phrases and single words can be tested.

\section{Fast Noun Phrase Parsing}
\label{em}
A fast and robust noun phrase parser is a key to the exploration of
syntactic phrase indexing. Noun phrase parsing, or noun phrase structure
analysis ( also known as compound noun analysis\footnote{Strictly
speaking, however, compound noun analysis is a special case of noun
phrase analysis, but the same technique can often be used for both.}),
is itself an important research issue in computational linguistics and
natural language processing.
Long noun phrases, especially long compound nouns such as ``{\em
information retrieval technique}'', 
generally
have ambiguous structures. For instance, ``{\em information retrieval
technique}'' 
has two possible structures: ``{\em [[information retrieval]
technique]}'' and ``{\em [information [retrieval technique]]}''.   
A principal difficulty in noun phrase structure analysis
 is to resolve such structural ambiguity. When a large corpus is
available, which is true for an IR task, statistical preference of word
combination or word modification can be a good clue for such
disambiguation. 
As summarized in \cite{Lauer2}, there are two different models for
corpus-based parsing of noun phrases: the adjacency model and the
dependency model. The difference between the two models can be
illustrated by the example compound 
noun  ``{\em information retrieval technique}''.  
In the adjacency model, the structure would be decided by looking at the
adjacency association of ``{\em information retrieval}'' and ``{\em
retrieval  technique}''. ``{\em information retrieval}'' will be grouped
first, if ``{\em information retrieval}'' has  a stronger association
than ``{\em retrieval technique}'', otherwise, 
``{\em retrieval technique}'' will be grouped first. In the dependency
model, however, the structure  would be decided
by  looking at the dependency between ``{\em information}'' and ``{\em
retrieval}'' (i.e.,
the tendency for ``{\em information}'' to modify ``{\em retrieval}'')
and the dependency
between ``{\em information}'' and ``{\em technique}''. If ``{\em
information}'' has a stronger
dependency association with ``{\em retrieval}'' than with ``{\em
technique}'', ``{\em information
retrieval}'' will be grouped first, otherwise, ``{\em retrieval
technique}'' will be grouped first. 
The adjacency model dates at least from  \cite{Marcus} and has been explored
recently in \cite{Liberman,Pustejovsky,Resnik,Lauer2,NYU-TREC3,ACL}; The
dependency model has mainly been studied in \cite{Lauer1}. 
Evans and Zhai \cite{ACL} use primarily the adjacency model, but the
association score also takes into account some degree of dependency.
Lauer \cite{Lauer2} compared the adjacency model and the dependency
model for compound noun disambiguation, and
concluded that the dependency model provides a substantial advantage
over the adjacency model. 

We now propose a probabilistic model in which
the dependency structure, or the modification structure, of a noun phrase is 
treated as ``hidden'', similar to the tree structure in the
 probabilistic context-free grammar \cite{PCFG}. The basic idea is as follows.

A noun phrase can be assumed to be generated from a 
word modification structure (i.e., a dependency structure). Since noun
phrases with more than two words are
structurally ambiguous, if we only observe the noun phrase, then the actual
structure that generates the noun phrase is  ``hidden''. We treat the
noun phrases with their possible structures as the  
complete data and the noun phrases occurring in the corpus (without
the structures) as the observed incomplete 
data. In the training phase, an Expectation Maximization (EM) algorithm
\cite{EM} can be used to estimate the 
parameters of word modification probabilities 
by iteratively maximizing the conditional expectation of the likelihood of the 
complete data given the observed incomplete data and a previous estimate
of the parameters. In the parsing phase, 
a noun phrase is assigned the structure that has the maximum conditional
probability given the noun phrase. 

Formally, assume that each noun phrase is generated using a word
modification structure. For example, ``{\em information retrieval
technique}'' may be generated using either the structure
``$[X_1 [X_2 X_3]]$'' or the structure ``$[[X_1 X_2] X_3]$''. The log
likelihood of generating a noun phrase, given the set of noun phrases
observed in a corpus $NP=\{np_i\}$ can be written as:
\[ L(\phi) = \sum_{np_i \in NP} c(np_i) log \sum_{s_j \in S}P_\phi(np_i,s_j) \]
where, $S$ is the set of all the possible modification structures; $c(np_i)$ is
the count of the noun phrase $np_i$ in the corpus; and 
$P_\phi(np_i,s_j)$ gives the probability of deriving the noun phrase
$np_i$ using the modification structure $s_j$. 

With the simplification that generating a noun phrase from a
modification structure is the same as generating all the corresponding
word modification pairs in the noun phrase and with the assumption that
each word modification pair in the noun phrase is generated
independently,
$P_\phi(np_i,s_j)$ can further be written as 
\[ P_\phi(np_i,s_j)=P_\phi(s_j)\prod_{(u, v) \in M(np_i,s_j)}
P_\phi(u,v)^{c(u,v;np_i,s_j)} \]
where, $M(np_i,s_j)$ is the set of all word pairs $(u,v)$ in $np_i$
such that
$u$ modifies (i.e., depends on) $v$ according to $s_j$.\footnote{For
example, if $np_i$ is ``{\em information retrieval technique}'', and
$s_j$ is ``$[[X_1 X_2] X_3]$'', then, $M(np_i,s_j)$ = \{($information$,
$retrieval$), ($retrieval$, $technique$)\}.}  
$c(u,v; np_i, s_j)$
is the count of the modification pairs $(u,v)$ being generated 
when $np_i$ is derived from $s_j$.
 $P_\phi(s_j)$ is the probability of structure $s_j$; while
$P_\phi(u,v)$ is the probability of generating the word pair $(u,v)$
given any word modification relation. $P_\phi(s_j)$ and $P_\phi(u,v)$
are subject
to the constraint of summing up to 1 over all modification structures and over
all possible word combinations respectively.\footnote{One problem with
such simplification is that the model may generate
a set of word modification pairs that do not form a noun phrase,
although such ``illegal noun phrases'' are never observed. A better
model  would be to write the probability of each word modification pair
as the conditional probability of the modifier (i.e., the modifying
word) given the head (i.e., the word being modified). That is,\\
\( P_\phi(np_i,s_j)= \) 
\[ \ \ \ \ P_\phi(s_j)P_\phi(h(np_i)|s_j)\prod_{(u, v) \in M(np_i,s_j)}
P_\phi(u|v)^{c(u,v;np_i,s_j)} \]
where $h(np_i)$ is the head (i.e., the last word) of the noun phrase
$np_i$\cite{Lafferty-Comm}.
}

The model is clearly a special case of the class of the algebraic
language models, in which the probabilities are expressed as polynomials
in the parameters\cite{Notes}. For such models, the M-step in the EM
algorithm can be carried
out exactly, and the parameter update formulas are:
\begin{eqnarray*}
\lefteqn{ P_{n+1}(u,v)}   \\
& & =\lambda_1^{-1}\sum_{np_i \in NP} c(np_i)\sum_{s_j \in S}P_{n}(s_j |
np_i)c(u,v;np_i,s_j)  \\
\ \\
\lefteqn{ P_{n+1}(s_k)} \\
&& =\lambda_2^{-1}\sum_{np_i \in NP} c(np_i)P_{n}(s_k | np_i) 
\end{eqnarray*}
where, $\lambda_1$ and $\lambda_2$ are the Lagrange multipliers corresponding
to the two constraints mentioned above, and are given by the following
formulas:
\small
\begin{eqnarray*}
\lambda_1 =\sum_{(u,v) \in WP} \sum_{np_i \in NP} c(np_i)\sum_{s_j \in
S}P_{n}(s_j | np_i)c(u,v;np_i,s_j)  \\
\end{eqnarray*}
\begin{eqnarray*}
\lambda_2 = \sum_{s_k \in S}\sum_{np_i \in NP} c(np_i)P_{n}(s_k | np_i) \\
\end{eqnarray*}
\normalsize
where, $WP$ is the set of all possible word pairs.\\
\ \\
$P_{n}(s_{j} | np_{i})$ can be computed as:
\small
\begin{eqnarray*}
\lefteqn{P_{n}(s_{j}|np_{i})} \\
&& = \frac{P_{n}(np_{i}, s_{j})}{P_{n}(np_{i})} \\
&& = \frac{P_{n}(np_{i}, s_{j})}{\sum_{s_k \in S}P_{n}(np_i,s_k)} \\
&& = \frac{P_{n}(s_j)\prod_{u, v \in M(np_i,s_j)}
P_{n}(u,v)^{c(u,v;np_i,s_j)} }{
\sum_{s_k \in S}(P_{n}(s_k)\prod_{u, v \in M(np_i,s_k)}
P_{n}(u,v)^{c(u,v;np_i,s_k)})}
\end{eqnarray*}
\normalsize

The EM algorithm ensures that $L(n+1)$ is greater than $L(n)$. In other words,
every step of parameter update increases the likelihood. Thus, at the time
of training, the parser can first randomly initialize the parameters, and
then, iteratively update the parameters according to the update formulas
until the increase of the likelihood is smaller than  some pre-set
threshold.\footnote{For the experiments reported in this paper, the
threshold is 2.}
In the implementation described here, the maximum length of any noun
phrase
is limited to six. In practice, this is not a very tight limit, since
simple noun phrases with more than six words are
quite rare. Summing over all the possible structures for any noun
phrase is computed by enumerating all the possible structures with an
equal length as the noun phrase. For example, in the case of a
three-word noun phrase, only two structures need to be enumerated.   

At the time of parsing noun phrases, the structure of any noun phrase
$np$ ($S(np)$) is
determined by 
\begin{eqnarray*}
S(np)&=&argmax_{s}P(s|np) \\
 &=&argmax_{s} P(np|s)P(s) \\
 &=&argmax_{s} \prod_{(u,v)\in M(np,s)}P(u,v)P(s) 
\end{eqnarray*}

We found that the parameters
 may easily be biased owing to data
sparseness. For example, the modification structure parameters naturally
prefer left association to right association
in the case of three-word noun phrases, when the data is sparse. Such bias
in the parameters of the modification structure probability will be propagated
to the word modification parameters when the parameters are iteratively
updated using EM algorithm. In the experiments reported in this paper, 
an over-simplified solution is adopted. We simply fixed the modification
structure parameter and assumed every 
dependency
structure is  equally likely. 

Fast training is achieved  by reading all the noun phrase instances into
memory.\footnote{An alternative way would
be to keep the corpus in the disk. In this way, it is not necessary to
split the corpus, unless it is extremely large.}  This forces us to
split the whole noun phrase corpus into 
small chunks for training. In the experiments reported in this paper, we
split the corpus into chunks of a size of around 4 megabytes.  Each
chunk has about 170,000 (or about 100,000 unique) raw multiple word noun
phrases. The parameters estimated
on each sub-corpus are then merged (averaged). 
We do not know how much the  merging of parameters affects the parameter
estimation,
but it seems that a majority of phrases are  
 correctly parsed with the merged parameter estimation, based on a rough
check of the parsing results. 
With this approach, it  takes a 133-MHz DEC Alpha workstation  about
5 hours to train the parser over the noun phrases from a 250-megabyte
text corpus. 
Parsing is much faster, taking less than 1 hour to parse
 all noun phrases in the corpus of a 250-megabyte text.
The parsing speed can be scaled up
to gigabytes of text, even when the parser needs to be re-trained over
the noun phrases in
the whole corpus. However, the speed has not taken into account
the time required for extracting the noun phrases for training. In the
experiments described in the following section, the CLARIT noun phrase
extractor is used to extract all the noun phrases from the
250-megabyte text corpus. 

After the training on each chunk, the estimation of the parameter of 
word modifications is smoothed to account for the unseen word 
modification pairs. Smoothing is made by ``dropping'' a certain 
number of parameters that have the least probabilities,
taking out the probabilities of the dropped parameters,  
and evenly distributing these probabilities
among all the unseen word pairs as well as those pairs of the dropped
parameters. It is unnecessary to keep the dropped parameters after
smoothing, thus this method of smoothing helps reduce the memory
overload when merging parameters.  In the experiments reported in the
paper,
nearly half of the total number of word pairs seen in the training chunk
were dropped. Since, word pairs with the least probabilities generally
occur quite
rarely in the corpus and usually represent semantically illegal word
combinations, dropping such word pairs does not affect the parsing
output so significantly as it seems. In fact, it may not affect the parsing
decisions for the majority of noun phrases in the corpus at all.

The potential parameter
space for the probabilistic model can be extremely large, when the size
of the training corpus is getting larger. 
One solution to this problem 
is to use a class-based model similar to the one proposed in
\cite{IBM-class} or use parameters of conceptual association rather than
word association, as
discussed in \cite{Lauer1}\cite{Lauer2}. 

\section{Experiment Design}
\label{exp}

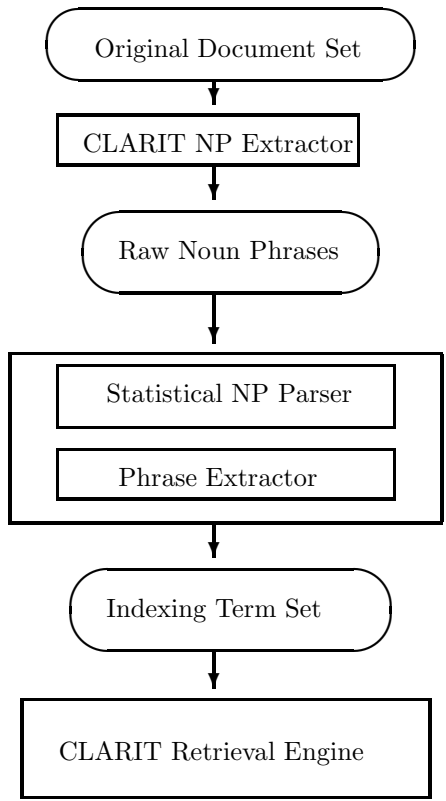
\begin{figure}[htb]
\setlength{\unitlength}{0.012500in}%
\begingroup\makeatletter
\def\x#1#2#3#4#5#6#7\relax{\def\x{#1#2#3#4#5#6}}%
\expandafter\x\fmtname xxxxxx\relax \def\y{splain}%
\ifx\x\y   
\gdef\SetFigFont#1#2#3{%
  \ifnum #1<17\tiny\else \ifnum #1<20\small\else
  \ifnum #1<24\normalsize\else \ifnum #1<29\large\else
  \ifnum #1<34\Large\else \ifnum #1<41\LARGE\else
     \huge\fi\fi\fi\fi\fi\fi
  \csname #3\endcsname}%
\else
\gdef\SetFigFont#1#2#3{\begingroup
  \count@#1\relax \ifnum 25<\count@\count@25\fi
  \def\x{\endgroup\@setsize\SetFigFont{#2pt}}%
  \expandafter\x
    \csname \romannumeral\the\count@ pt\expandafter\endcsname
    \csname @\romannumeral\the\count@ pt\endcsname
  \csname #3\endcsname}%
\fi
\endgroup
\begin{picture}(180,330)(60,475)
\thicklines
\put(152,703){\oval(124,34)}
\put(152,553){\oval(134,34)}
\put( 80,740){\framebox(125,20){}}
\put( 80,630){\framebox(140,25){}}
\put( 80,600){\framebox(140,20){}}
\put( 60,590){\framebox(180,70){}}
\put( 65,475){\framebox(170,40){}}
\put(145,775){\vector( 0,-1){ 10}}
\put(145,740){\vector( 0,-1){ 15}}
\put(152,790){\oval(154,30)}
\put(145,685){\vector( 0,-1){ 20}}
\put( 80,490){\makebox(0,0)[lb]{\smash{\SetFigFont{10}{14.4}{rm}CLARIT
Retrieval Engine }}}
\put(145,590){\vector( 0,-1){ 15}}
\put(145,535){\vector( 0,-1){ 15}}
\put( 95,785){\makebox(0,0)[lb]{\smash{\SetFigFont{10}{14.4}{rm}Original
Document Set}}}
\put( 90,745){\makebox(0,0)[lb]{\smash{\SetFigFont{10}{14.4}{rm}CLARIT
NP Extractor}}}
\put(105,700){\makebox(0,0)[lb]{\smash{\SetFigFont{10}{14.4}{rm}Raw Noun
Phrases}}}
\put(100,640){\makebox(0,0)[lb]{\smash{\SetFigFont{10}{14.4}{rm}Statistical NP 
Parser}}}
\put(105,605){\makebox(0,0)[lb]{\smash{\SetFigFont{10}{14.4}{rm}Phrase
Extractor}}}
\put(100,550){\makebox(0,0)[lb]{\smash{\SetFigFont{10}{14.4}{rm}Indexing
Term Set}}}
\end{picture}
\caption{Phrase indexing experiment procedure}

\label{procedure}
\end{figure}

We used the CLARIT commercial retrieval system as a retrieval engine  to 
test the effectiveness of  different indexing sets. The CLARIT system
uses the vector space retrieval model\cite{Salton}, in which documents
and the query are all represented by a vector of
weighted terms (either single words or phrases), and the relevancy
judgment is based
on the similarity (measured by the cosine measure) between the query
vector and any document
vector\cite{CLARIT-TREC1,CLARIT-TREC2,CLARIT-TREC4}. The experiment
procedure 
is described 
by Figure~\ref{procedure}.
 
First, the original database is parsed to form different 
sets of indexing terms (say, using  different combination of phrases).
Then, each indexing set is passed
to the CLARIT retrieval engine as a source document set. 
The CLARIT system is configured to accept the indexing set
we passed as is to ensure that the actual indexing terms
used inside the CLARIT system are exactly those generated.

It is possible to generate three different kinds/levels of indexing
units from a noun phrase:  (1) single words; (2) head modifier pairs
(i.e., any word pair in the noun phrase that has a linguistic
modification relation);
and (3) the full noun phrase.   For example, from the
phrase structure 
``[[[heavy=construction]=industry]]=group]'' (a real example from WSJ90), 
it is
possible to generate the following candidate terms:
\begin{verbatim}
SINGLE WORDs:
  heavy, construction, industry, group 
HEAD MODIFIERS:
  construction industry, industry group, 
  heavy construction 
FULL NP:
  heavy construction industry group 
\end{verbatim}

Different combinations of the three kinds of terms can be selected for
indexing.
In particular, the indexing set formed solely of single words is used as
a baseline to test the effect of using  phrases. In the experiments reported 
here, we generated four different
combinations of phrases:

\begin{verbatim}
-- WD-SET: 
  single word only (no phrases, baseline)
-- WD-HM-SET: 
  single word + head modifier pair 
-- WD-NP-SET: 
  single word + full NP
-- WD-HM-NP-SET: 
  single word + head modifier + full NP
\end{verbatim}

The results from these different phrase sets are discussed in the next section.

\section{Results analysis}
\label{res}

We used, as our document set, the Wall Street Journal database in
Tipster 
Disk2 \cite{TREC5} the size of which
is about 250 megabytes. We performed the experiments by using
the TREC-5 ad hoc topics (i.e., TREC topics 251-300). Each run
involves an automatic feedback with the top 10 documents returned
from the initial retrieval. The CLARIT automatic feedback is performed by
adding terms from a query-specific
thesaurus extracted from the top N documents returned from the initial
retrieval\cite{CLARIT-TREC2}.  
The results are evaluated using the standard measures of recall and
precision. Recall measures how many of the relevant documents have 
actually been
retrieved. Precision measures how many of the retrieved documents are indeed
relevant. They are calculated by the following simple formulas:

\small
\[Recall = \frac{number\ of\ relevant\ items\ retrieved}{total\ number\
of\ relevant\ items\ in\ collection} \]
\[Precision = \frac{number\ of\ relevant\ items\ retrieved}{total\
number\ of\  items\ retrieved} \]

\normalsize
 We used the standard TREC evaluation package provided by Cornell 
University and used the judged-relevant documents from the TREC evaluations
as the gold standard\cite{TREC-Eval}.  

In  Table~\ref{trec5-fb}, we give
a summary of the results and compare  the three
 phrase combination runs with the corresponding baseline run.
In the table, ``Ret-rel'' means ``retrieved-relevant'' and refers to the
total number of relevant
documents retrieved. ``Init Prec'' means ``initial precision'' and refers
to the highest level of precision over all the points of recall. ``Avg
Prec'' means ``average
precision'' and is the average of all the precision values computed after
each new 
relevant document is retrieved. 

 It is clear that phrases help both recall and precision when
supplementing single words, as can be seen from the improvement of 
all phrase runs (WD-HM-SET, WD-NP-SET, WD-HM-NP-SET) over the
single word run WD-SET. 

\tiny
\begin{table}[hbt]
\begin{center}
\begin{tabular}{|r|c|c|c|}
\hline
Experiments   & Recall (Ret-Rel) & Init Prec & Avg Prec \\ \hline
WD-SET  & 0.56(597) & 0.4546 & 0.2208 \\ \hline
WD-HM-SET & 0.60( 638 ) & 0.5162 & 0.2402 \\
inc over WD-SET &  7\% &  14\% & 9\% \\ \hline
WD-NP-SET & 0.58(613)  & 0.5373 & 0.2564 \\
inc over WD-SET   &4\% & 18\% & 16\% \\ \hline
WD-HM-NP-SET &0.63(666) & 0.4747 & 0.2285 \\ 

inc over WD-SET & 13\% & 4\% &  3\% \\ \hline
\multicolumn{4}{|c|}{Total relevant documents: 1064}\\ \hline
\end{tabular}
\protect\caption{Effects of Phrases with feedback and TREC-5 topics}
\label{trec5-fb}
\end{center}
\end{table}

\normalsize
It can also be seen that when only one kind of phrase (either
the full NPs or the head modifiers) is used to supplement
the single words, each can lead to a great improvement in 
precision. However, when we combine the two
kinds of phrases, the effect is a greater improvement in
recall rather than  precision. The fact that each kind of phrase
can improve precision significantly when used separately shows
that these phrases are indeed very useful for indexing. 
The combination of phrases results in only a smaller precision improvement
but causes a much greater increase in recall. This may indicate that
more experiments
are needed to understand how to combine and weight 
different phrases effectively. 

The same parsing method has also been used to generate
phrases from the same data for the CLARIT NLP track experiments in
TREC-5\cite{TREC5-NLP}, and similar results were obtained, although the
WD-NP-SET
was not tested. The results in \cite{TREC5-NLP} are not identical to the
results here, because they are based on two separate training processes.
It is possible that different
training processes may result in slightly different parameter
estimations,  because the corpus is arbitrarily segmented into chunks of
only roughly 4 megabytes for training, and the chunks actually used in
different 
training processes may vary slightly.

\section{Conclusions}
\label{con}

Information retrieval provides a good way to quantitatively (although
indirectly) evaluate various
NLP techniques. 
We explored the application of  a fast statistical noun phrase parser to
enhance document indexing in information retrieval. We proposed a new
probabilistic model for 
noun phrase parsing and developed a fast noun phrase parser that can
handle relatively large amounts of text efficiently. The effectiveness
of enhancing document indexing with the syntactic phrases provided by 
the noun phrase parser was evaluated  on the Wall Street Journal
database in Tipster Disk2 using
50 TREC-5 ad hoc topics.   
Experiment results on
this 250-megabyte document collection have shown that using different
kinds of syntactic 
phrases provided by the noun phrase parser to supplement single words
for indexing  can significantly improve
the retrieval performance, which is more encouraging than many early
experiments on syntactic phrase indexing.  Thus, using selective NLP,
such as the noun phrase parsing
technique we proposed, is not only feasible for use in information
retrieval, but also effective in enhancing the retrieval
performance.\footnote{Whether such syntactic phrases are more effective
than simple statistical phrases (e.g., high frequency word bigrams)
remains to be tested.}

There are two lines of future work:

First, the  results from information retrieval experiments often show variances
on different kinds of document collections and different sizes of
collections.  It is thus desirable to test the noun
phrase parsing technique in other and larger collections. More experiments and 
analyses are also needed to  better understand how to more effectively 
combine different 
phrases  with 
single words. In addition,  
it is very important to study how such phrase effects interact with other
useful IR techniques such as relevancy feedback, query expansion, and term 
weighting. 

Second, it is desirable to study how the parsing quality (e.g.,  in terms of
the ratio of phrases  parsed correctly) would affect the retrieval 
performance. It is very interesting to try the conditional probability
model as mentioned in
a footnote in section~\ref{em}  The
improvement of the probabilistic model of noun phrase parsing may result
in phrases of
higher quality than the phrases produced by the current noun phrase parser. 
Intuitively, the use of 
higher quality phrases might enhance document indexing more effectively, but 
this again needs to be tested.

\section{Acknowledgments}

The author is especially grateful to David A. Evans for his advising and
supporting of this work. Thanks are also due to John Lafferty,
Nata{\v{s}}a Mili{\'{c}}-Frayling, Xiang Tong, and two anonymous
reviewers  for their useful comments. Naturally, the author alone is 
responsible
for all the errors.

\end{document}